\documentclass{article}
\usepackage{ijcai23}

\usepackage{times}
\usepackage{soul}
\usepackage{url}
\usepackage[hidelinks]{hyperref}
\usepackage[utf8]{inputenc}
\usepackage[small]{caption}
\usepackage{graphicx}
\usepackage{amsmath}
\usepackage{amsthm}
\usepackage{booktabs}
\usepackage{algorithm}
\usepackage{algorithmic}
\usepackage[switch]{lineno}

\usepackage{amssymb,amsfonts}
\usepackage{textcomp}
\usepackage{multirow}
\usepackage{newfloat}
\usepackage{verbatim}

\title{CostFormer:Cost Transformer for Cost Aggregation in Multi-view Stereo}

\author{
Weitao Chen\footnotemark[1]$^1$
\and
Hongbin Xu\footnotemark[1]$^{1,2}$\and
Zhipeng Zhou$^{1}$\and
Yang Liu$^1$\and
Baigui Sun\footnotemark[2]$^1$\and
Wenxiong Kang\footnotemark[2]$^2$\and
Xuansong Xie$^1$
\affiliations
$^1$Alibaba Group\\
$^2$South China University of Technology\\
\emails
\{hillskyxm, hongbinxu1013\}@gmail.com,
baigui.sbg@alibaba-inc.com,
auwxkang@scut.edu.cn
}

\begin{document}

\maketitle

\renewcommand{\thefootnote}{\fnsymbol{footnote}} 
\footnotetext[1]{These authors contributed equally to this work.} 
\footnotetext[2]{Corresponding authors.} 

\begin{abstract}
    The core of Multi-view Stereo(MVS) is the matching process among reference and source pixels. Cost aggregation plays a significant role in this process, while previous methods focus on handling it via CNNs. This may inherit the natural limitation of CNNs that fail to discriminate repetitive or incorrect matches due to limited local receptive fields. To handle the issue, we aim to involve Transformer into cost aggregation. However, another problem may occur due to the quadratically growing computational complexity caused by Transformer, resulting in memory overflow and inference latency. In this paper, we overcome these limits with an efficient Transformer-based cost aggregation network, namely CostFormer. The Residual Depth-Aware Cost Transformer(RDACT) is proposed to aggregate long-range features on cost volume via self-attention mechanisms along the depth and spatial dimensions. Furthermore, Residual Regression Transformer(RRT) is proposed to enhance spatial attention. The proposed method is a universal plug-in to improve learning-based MVS methods.
\end{abstract}

\section{Introduction}

Given a series of calibrated images from different views in one scene, Multi-view Stereo (MVS) aims to recover the 3D information of the observed scene.
It is a fundamental problem in computer vision and widely applied to robot navigation, autonomous driving, augmented reality, and etc.
Recent learning-based MVS networks  \cite{yao2018mvsnet,gu2020cascade,wang2021patchmatchnet} have achieved inspiring success both in the quality and the efficiency of 3D reconstruction. 
Generally, deep MVS approaches consist of the following five steps: feature extraction from multi-view images via CNN network with shared weights, differentiable warping to align all source features to the reference view, matching cost computation from reference features and aligned source features, matching cost aggregation or regularization, depth or disparity regression.  

\begin{figure}[t!]
	\center
        \includegraphics[width=\hsize]{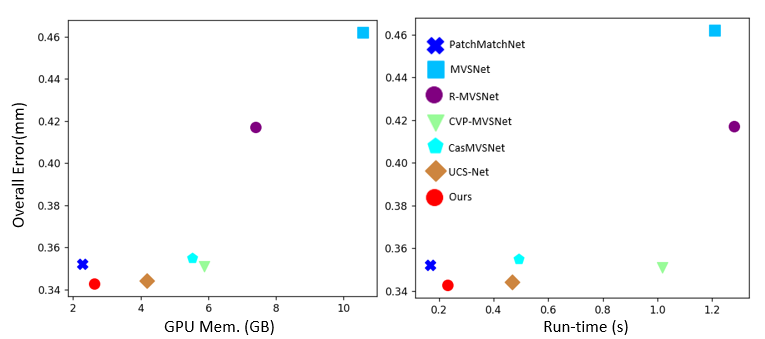}
	\caption{Comparison with state-of-the-art MVS methods on DTU. Relationship between error, GPU memory and run-time with image size 1152×864.}
	\label{fig:compar}
\end{figure}

Current progresses in learning-based MVS primarily concentrate on the limitation of reconstruction quality \cite{wei2021aa,yang2020cost}, memory consumption  \cite{yan2020dense,wei2021aa}, and efficiency \cite{wang2021patchmatchnet,wang2021itermvs}.
The basic network architecture of these works is based on the pioneering backbone network called MVSNet \cite{yao2018mvsnet}, which provides an elegant and stable baseline.
However, instead of taking the inheritance of network design principle in MVSNet \cite{yao2018mvsnet} for granted, we can rethink the task of MVS problem as a dense correspondence problem \cite{hosni2012fast} alternatively.
The core of MVS is a dense pixelwise correspondence estimation problem that searches the corresponding pixel of a specific pixel in the reference image along the epipolar line in all warped source images.
No matter which task this correspondence estimation problem is applied to, the matching task can be boiled down to a classical matching pipeline \cite{scharstein2002taxonomy}: (1) feature extraction, and (2) cost aggregation.
In learning-based MVS methods, the transition from traditional hand-crafted features to CNN-based features inherently solves the former step of the classical matching pipeline via providing powerful feature representation learned from large-scale data.
However, \emph{handling the cost aggregation step by matching similarities between features without any prior usually suffers from the challenges due to ambiguities generated by repetitive patterns or background clutters} \cite{cho2021cats}.
Consequently, a typical solution in MVSNet and its variants \cite{yao2018mvsnet,gu2020cascade,wang2021patchmatchnet} is to apply a 3D CNN or an RNN to regularize the cost volume among reference and source views, rather than directly rely on the quality of the initial correlation clues in cost volume.
Although formulated variously in previous methods, these methods either use hand-crafted techniques that are agnostic to severe deformations or inherit the limitation of CNNs, e.g. limited receptive fields, unable to discriminate incorrect matches that are locally consistent.

In this work, we focus on the \emph{cost aggregation step} of cost volume and propose a novel cost aggregation Transformer (\textbf{CostFormer}) to tackle the issues above. Our CostFormer is based on Transformer \cite{vaswani2017attention}, which is renowned for its global receptive field and long-range dependent representation. By aggregating the matching cost in the cost volume, our aggregation network can explore global correspondences and refine the ambiguous matching points effectively with the help of the self-attention (SA) mechanism in Transformer.
Though the promising performances of Vision Transformers have been proven in many applications \cite{dosovitskiy2020image,sun2021loftr}, the time and memory complexity of the key-query dot product interaction in conventional SA grow quadratically with the spatial resolution of inputs.
Hence, replacing 3D CNN with Transformer may result in unexpected extra occupancy in memory and latency in inference.
Inspired by  \cite{wang2021patchmatchnet}, we further introduce the Transformer architecture into an iterative multi-scale learnable PatchMatch pipeline.
It inherits the advantages of the long-range receptive field in Transformers, improving the reconstruction performance substantially.
Meantime, it also maintains a balanced trade-off between efficiency and performance, which is competitive in the inference speed and parameters magnitude compared with other methods.

Our main contributions are as follows: 

(1) In this paper, we propose a novel Transformer-based cost aggregation network called CostFormer, which
can be plugged into learning-based MVS methods to improve cost volume effectively.
(2) CostFormer applies an efficient Residual Depth-Aware Cost Transformer to cost volume, extending 2D spatial attention to 3D depth and spatial attention. 
(3) CostFormer applies an efficient Residual Regression Transformer between cost aggregation and depth regression, keeping spatial attention.  
(4) The proposed CostFormer brings benefits to learning-based MVS methods when evaluating DTU \cite{journals/ijcv/AanaesJVTD16}, Tanks $\&$ Temples \cite{journals/tog/KnapitschPZK17} ETH3D \cite{conf/cvpr/SchopsSGSSPG17} and BlendedMVS \cite{yao2020blendedmvs} datasets. 

\section{Related Work}
\label{sec_related_work}

\subsection{Learning-based MVS Methods}

Powered by the great success of deep learning-based techniques, many learning-based methods have been proposed to boost the performance of Multi-view Stereo.
MVSNet  \cite{yao2018mvsnet} is a landmark for the end-to-end network that infers the depth map on each reference view for the MVS task.
Feature maps extracted by a 2D CNN on each view are reprojected to the same reference view to build a variance-based cost volume.
A 3D CNN is further used to regress the depth map.
Following this pioneering work, lots of efforts have been devoted to boosting speed and reducing memory occupation.
To relieve the burden of huge memory cost, recurrent neural networks are utilized to regularize the cost volume in AA-RMVSNet  \cite{wei2021aa}.
Following a coarse-to-fine manner to develop a computationally efficient network, a recent strand of works divide the single cost volume into several cost volumes at multiple stages, like CasMVSNet  \cite{gu2020cascade}, CVP-MVSNet  \cite{yang2020cost}, UCSNet  \cite{conf/cvpr/ChengXZLLRS20}, and etc. 
Inspired by the traditional PatchMatch stereo algorithm, PatchMatchNet  \cite{wang2021patchmatchnet} inherits the pipeline in PatchMatch stereo in an iterative manner and extend it into a learning-based end-to-end network.

\begin{figure*}[t]
	\center
        \includegraphics[width=\hsize]{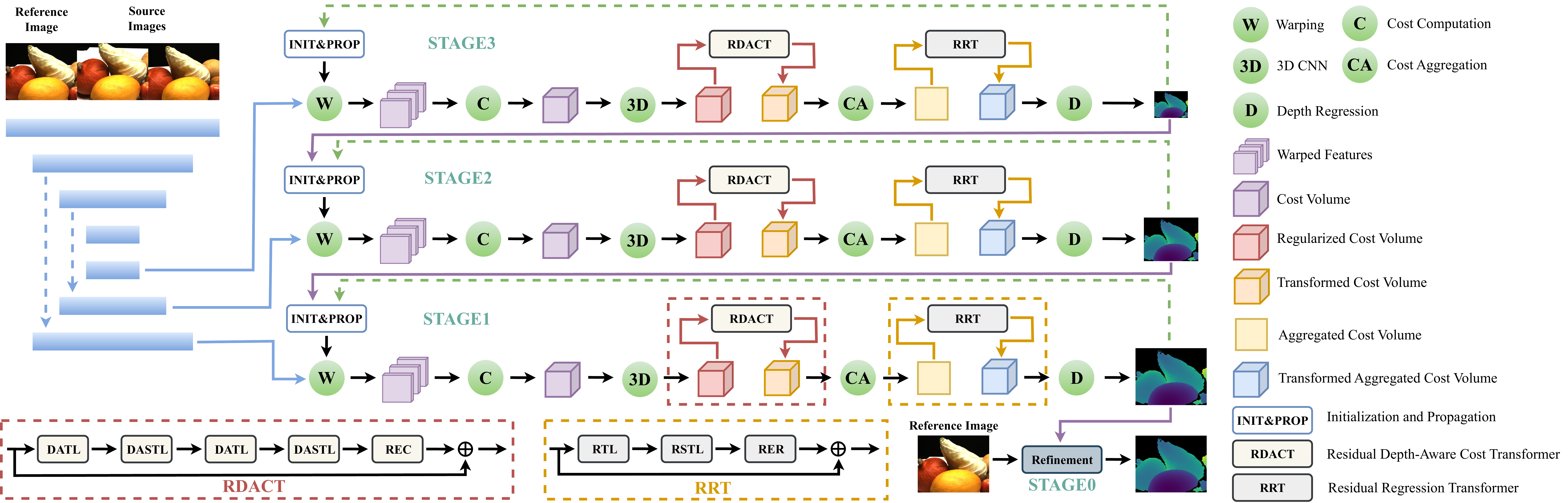}
	\caption{Structure of CostFormer based on PatchMatchNet.}
	\label{fig:CostFormer}
\end{figure*}

\subsection{Vision Transformer}

The success of Transformer  \cite{vaswani2017attention} and its variants  \cite{dosovitskiy2020image,liu2021swin} have motivated the development of Neural Language Processing in recent years.
Borrowing inspiration from these works, Transformer has been successfully extended to vision tasks and proven to boost the performance of image classification  \cite{dosovitskiy2020image}.
Following the pioneering work, many efforts are devoted to boosting the development of various vision tasks with the powerful representation ability of Transformer.

In  \cite{li2021revisiting}, the application of Transformer in the classic stereo disparity estimation task is investigated thoughtfully.
Swin Transformer  \cite{liu2021swin} involves the hierarchical structure into Vision Transformers and computes the representation with shifted windows.
Considering Transformer's superiority in extracting global content information via attention mechanism, many works attempt to utilize it in the task of feature matching.
Given a pair of images, CATs  \cite{cho2021cats} explore global consensus among correlation maps extracted from a Transformer, which can fully leverage the self-attention mechanism and model long-range dependencies among pixels.
LoFTR  \cite{sun2021loftr} also leverages Transformers with a coarse-to-fine manner to model dense correspondence.
STTR  \cite{li2021revisiting} extends the feature matching Transformer architecture to the task of stereo depth estimation task in a sequence-to-sequence matching perspective.
TransMVSNet  \cite{ding2021transmvsnet} is the most relevant concurrent work compared with ours, which utilizes a Feature Matching Transformer (FMT) to leverage self-attention and cross-attention to aggregate long-range context information within and across images.
Specifically, the focus of TransMVSNet is on the enhancement of feature extraction before cost aggregation, while our proposed CostFormer aims to improve the cost aggregation process on cost volume.

\section{Methodology}
\label{sec_methodology}
In this section, we introduce the detailed architecture of the proposed CostFormer which focuses on the cost aggregation step of cost volume. 
CostFormer contains two specially designed modules called Residual-Depth Aware Cost Transformer (RDACT) and Residual Regression Transformer (RRT), which are utilized to explore the relation between pixels within a long range and the relation between different depth hypotheses during the evaluation process. In Section Preliminary, we give a brief preliminary on the pipeline of our method. Then we show the construction of RDACT and RRT respectively.  Finally, we show experiments.

\begin{figure*}[ht!]
\centering
\includegraphics[width=0.9\textwidth]{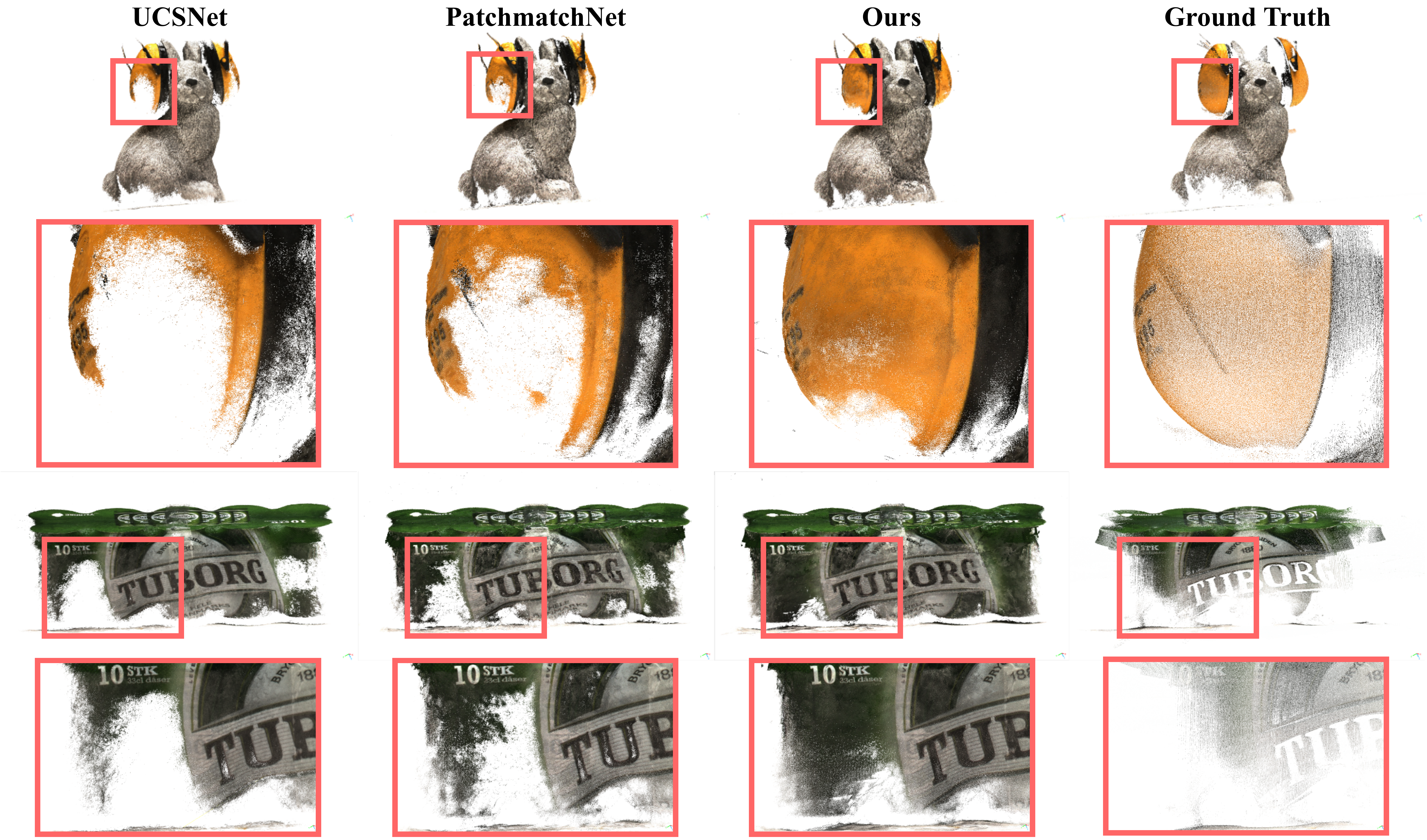}
\caption{Comparison of different methods on the DTU evaluation set. The backbone of CostFormer is PatchMatchNet here.}
\label{fig:dtu_comparison}
\end{figure*} 

\subsection{Preliminary}\label{PatchMatch-Review}

In general, the proposed RDACT and RRT can be integrated with arbitrary cost volume of learning-based MVS networks.  
Based on the patch match architecture \cite{wang2021patchmatchnet}, we further explore the issue of cost aggregation on cost volume. 
As shown in Figure \ref{fig:CostFormer}, CostFormer based on PatchMatchNet \cite{wang2021patchmatchnet} extracts feature maps from multi-view images and performs initialization and propagation to warp the features maps in source views to reference view. Given a pixel $p$ at the reference view and its corresponding pixel $p_{i,j}$ at the $i$-th source view under the $j$-th depth hypothesis $d_{j}$ is defined as:
        \begin{equation}p_{i,j}=K_{i}\cdot(R_{0,i}\cdot(K_{0}^{-1}\cdot p\cdot d_{j}) + t_{0,i})\end{equation}
where $R_{0,i}$ and $t_{0,i}$ denote the rotation and translation between the reference view and $i$-th source view.
$K_{0}$ and $K_{i}$ are the intrinsic matrices of the reference and $i$-th source view.
The warped feature maps at the $i$-th source view $F_{i}(p_{i,j})$ are bilinearly interpolated to remain the original resolution.
Then, a cost volume is constructed from the similarity of feature maps, and 3D CNNs are applied to regularize the cost volume.
Warped features from all source views are integrated into a single cost for each pixel $p$ and depth
hypothesis $d_{j}$ by computing the cost per hypothesis $S_i(p,j)^{g}$ via group-wise correction as follows: 
                \begin{equation}S_{i}(p,j)^{g} = \frac{G}{C}<F_{0}(p)^{g}, F_{i}(p_{i,j})^{g}> \in \mathbb{R}^{G}\end{equation}
where $G$ is the group number, $C$ is the channel number, $<\cdot, \cdot>$ is the inner product, $F_{0}(p)^{g}$ and $F_{i}(p_{i,j})^{g}$ are grouped reference feature map and grouped source feature map at the $i$-th view respectively.
Then they aggregate over the views with a pixel-wise view weight $w_{i}(p)$ to get $\overline S(p, j)$.

Taking no account of Transformer at the cost aggregation (CA) step, a CA module firstly utilizes a small network with 3D convolution with $1\times1\times1$ kernels to obtain a single cost, $\mathcal{C}$ $\in R^{H\times W\times D}$.
For a spatial window of $K_{e}$ pixels $\{p_{k}\}_{k=1}^{K_{e}}$ can be organized as a grid, per pixel additional offsets $\{ \Delta p_{k}\}_{k=1}^{K_{e}}$ can be learned for spatial adaptation.
The aggregated spatial cost $\widetilde {\mathcal{C}}(p, j)$ is defined as:
         \begin{equation}\widetilde{\mathcal{C}}(p, j) =\frac{1}{\sum_{k=1}^{K_{e}}w_{k}d_{k}}\overset{K_{e}}{\underset{k=1}{\sum}}w_{k}d_{k}\mathcal{C}(p+p_{k}+\Delta p_{k}, j)\end{equation}
where $w_{k}$ and $d_{k}$ weight the cost $\mathcal{C}$ based on feature and depth similarity.
Given the sampling positions ${(p+p_k+\Delta{p_k})}_{k=1}^{K_{e}}$, corresponding features from $F_0$ are extracted via bilinear interpolation. Then group-wise correlation is applied between the features at each sampling location and p. The results are concatenated into a volume on which 3D convolution layers with 1×1×1 kernels and sigmoid non-linearities are applied to output normalized weights ${\{w_k\}}_{k=1}^{K_{e}}$. The absolute difference in inverse depth between each sampling point and pixel p with their j-th hypotheses are collected. Then a sigmoid function on the inverted differences is applied to obtain ${\{d_k\}}_{k=1}^{K_{e}}$. 

The remarkable thing is that such cost aggregation inevitably suffers from challenges due to ambiguities generated by repetitive patterns or background clutters. The local mechanisms in ambiguities exist in many operations, such as local propagation and spatial adaptation by small learnable slight offset.
CostFormer significantly alleviates these problems through RDACT and RRT. The original CA module is also repositioned between RDACT and RRT.

After RRT, soft argmin is applied to get the regressed depth. Finally, a depth refinement module is designed to refine the depth regression. 

For CascadeMVS and other cascade architectures, CostFormer can be plugged into similarly.

\subsection{Residual Depth-Aware Cost Transformer}\label{RDACT}
In this section, we explore the details of the Residual Depth-Aware Cost Transformer (RDACT).
Each RDACT consists of two parts.
The first part is a stack of Depth-Aware Transformer layer (DATL) and Depth-Aware Shifted Transformer layer (DASTL), which deal with the cost volumes to explore the relations sufficiently.
The second part is the Re-Embedding Cost layer (REC) which recovers the cost volume from the first part.

Given a cost volume $\mathcal{C}_{0} \in \mathbb{R}^{H\times W\times D\times G}$, temporary intermediate cost volumes $\mathcal{C}_{1}$,$\mathcal{C}_{2}$,...,$\mathcal{C}_{L} \in \mathbb{R}^{H\times W\times D\times E}$ are firstly extracted by DATL and DASTL alternatively:
        \begin{equation}\mathcal{C}_{k} = \text{DASTL}_{k}(\text{DATL}_{k}(\mathcal{C}_{k-1})), k=1,2,...,L\end{equation}
where $\text{DATL}_{k}$ is the $k$-th Depth-Aware Transformer layer with regular windows, $\text{DASTL}_{k}$ is the $k$-th Depth-Aware Transformer layer with shifted windows, $E$ is the embedding dimension number of $\text{DATL}_{k}$ and $\text{DASTL}_{k}$. 

Then a Re-Embedding Cost layer is applied to the last $\mathcal{C}_{k}$, namely $\mathcal{C}_{L}$,  to recover $G$ from $E$.
The output of RDACT is formulated as:
                       \begin{equation}\mathcal{C}_{out} = \text{REC}(\mathcal{C}_{L}) + \mathcal{C}_{0}\end{equation}
where REC is the Re-Embedding Cost layer, and it can be a 3D convolution with $G$ output channels. If $E=G$, $\mathcal{C}_{out}$ can be simply formulated as:
                       \begin{equation}\mathcal{C}_{out} = \mathcal{C}_{L} + \mathcal{C}_{0}\end{equation}
This residual connection allows the aggregation of different levels of cost volumes; $\mathcal{C}_{out}$ instead of $\mathcal{C}_{0}$ is then aggregated by the original aggregation network described in section \ref{PatchMatch-Review}. The whole RDACT is shown in the red window in Figure \ref{fig:CostFormer}.            
\begin{figure*}[ht!]
\centering
\includegraphics[width=0.9\textwidth]{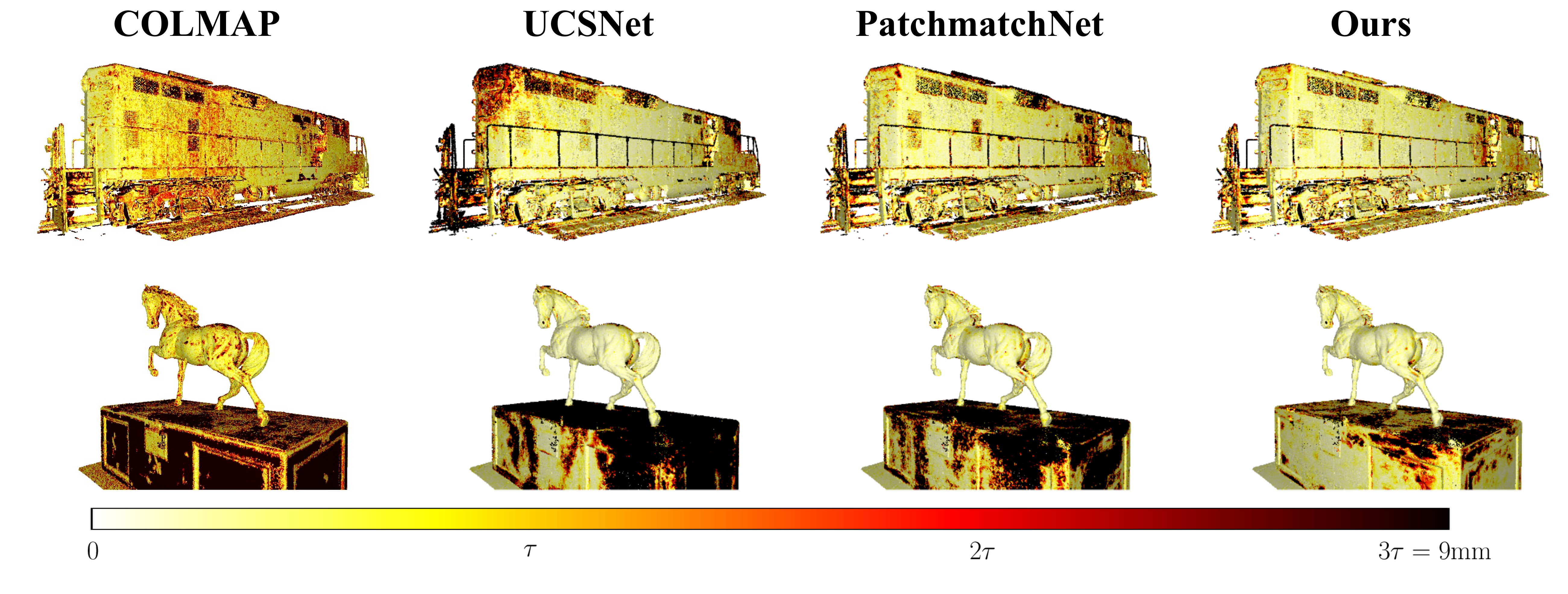}
\caption{ Comparison of different methods on Tanks\&Temples. The Recall reported by official benchmark is presented. }
\label{fig:tanks_comparison}
\end{figure*} 

Before introducing the construction of DATL and DASTL, we dive into the details of core constitutions called Depth-Aware Multi-Head Self-Attention (DA-MSA) and Depth-Aware Shifted Multi-Head Self-Attention (DAS-MSA).
Both DA-MSA and DAS-MSA are based on Depth-Aware Self-Attention Mechanism.
In order to explain Depth-Aware Self-Attention Mechanism, we supply the knowledge about Depth-Aware Patch Embedding and Depth-Aware Windows as preliminary. 

\noindent \textbf{Depth-Aware Patch Embedding}: 
Obviously, directly applying the attention mechanism for feature maps at pixel-wise level is quite costly in terms of GPU memory usage.
In order to tackle this issue, we propose a Depth-Aware Patch Embedding to reduce the high memory cost and get an additional regularization.
Specifically, 
given a grouped cost volume before aggregation $\mathcal{C}$ $\in \mathbb{R}^{H\times W\times D \times G}$, a depth-aware patch embedding is firstly applied to $\mathcal{C}$ to get tokens.
It consists of a 3D convolution with kernel size $h\times w\times d$ and a layer normalization.
To downsample the spatial sizes of cost volume and keep the depth hypotheses, we set $h$ and $w$ to more than 1 and $d$ as 1. So the sample ratio is adaptive for memory cost and run time.  
Before convolution, cost volume will be padded to fit the spatial sizes and downsampling ratio. After layer normalization(LN), these embedded patches are further partitioned by depth-aware windows. 

\noindent \textbf{Depth-Aware Windows}: Beyond the nonlinear and linear global self-attention, local self-attention within a window has been proven to be more effective and efficient.
As an example of 2D windows, Swin Transformer  \cite{liu2021swin} directly applies multi-head self-attention mechanisms on non-overlapping 2D windows to avoid the big computation complexity of global tokens.
Extended from the 2D spatial window, an embedded cost volume patch $\in \mathbb{R}^{H^{*}\times W^{*}\times D^{*}\times G}$ with depth information is partitioned into non-overlapping 3D windows.
These local windows are then transposed and reshaped to local cost tokens.
Assuming the sizes of these windows are $hs\times ws\times ds$, the total number of tokens is $\lceil{\frac{H^{*}}{h_{s}}}\rceil\times \lceil{\frac{W^{*}}{w_{s}}}\rceil\times \lceil{\frac{D^{*}}{d_{s}}}\rceil$.
These local tokens are further processed by the multi-head self-attention mechanism.


\noindent \textbf{Depth-Aware Self-Attention Mechanism}:
For a cost window token $X \in \mathbb{R}^{h_{s} \times w_{s} \times d_{s} \times G}$, the query, key, and value matrices $Q$, $K$ and $V \in \mathbb{R}^{h_{s} \times w_{s} \times d_{s} \times G}$ are computed as:
           \begin{equation}Q = XP_{Q}, K = XP_{K}, V = XP_{V}\end{equation}
where $P_{Q}$, $P_{K}$, and $P_{V} \in \mathbb{R}^{G \times G}$ are projection matrices shared across different windows.
By introducing depth and spatial aware relative position bias $B1 \in \mathbb{R}^{(h_{s} \times h_{s})\times (w_{s} \times w_{s})\times (d_{s} \times d_{s})}$ for each head, the depth-aware self-attention(DA-SA1) matrix within a 3D local window is thus computed as:
            \begin{scriptsize}\begin{equation}\text{DA-SA1}=Attention1(Q1, K1, V1) = SoftMax(\frac{Q1K1^{T}}{\sqrt{G}} + B1)V1\end{equation}\end{scriptsize}
Where $Q1$, $K1$ and $V1 \in \mathbb{R}^{h_{s}w_{s}d_{s} \times G}$ are reshaped from $Q$, $K$ and $V \in \mathbb{R}^{h_{s} \times w_{s} \times d_{s} \times G}$.  The process of DATL with
LayerNorm(LN) and multi-head DA-SA1 at the current level is formulated as:
             \begin{equation}\widehat X^{l} = \text{DA-MSA1}((\text{LN}(X^{l-1})) + X^{l-1}\end{equation}
By introducing depth-aware relative position bias $B2 \in \mathbb{R}^{d_{s} \times d_{s}}$ for each head, the depth-aware self-attention(DA-SA2) matrix along the depth dimension is an alternative module to DATL and thus computed as:
          \begin{scriptsize}\begin{equation}\text{DA-SA2}=Attention2(Q2, K2, V2) = SoftMax(\frac{Q2K2^{T}}{\sqrt{G}} + B2)V2 \end{equation}\end{scriptsize}
Where $Q2$, $K2$ and $V2 \in \mathbb{R}^{h_{s}w_{s} \times d_{s} \times G}$ are reshaped from $Q$, $K$ and $V \in \mathbb{R}^{h_{s} \times w_{s} \times d_{s} \times G}$. 
$B1$ and $B2$ will be along the depth dimension and lie in the range of $[-d_{s}+1, d_{s}-1]$.
Along the height and width dimension, $B1$ lies in the range of $[-h_{s}+1, h_{s}-1]$ and $[-w_{s}+1, w_{s}-1]$.
In practice, we parameterize a smaller-sized bias matrix $\overline{B1} \in \mathbb{R}^{(2h_{s}-1)\times (2w_{s}-1)\times (2d_{s}-1)}$ from $B1$ and perform the attention function
for $f$ times in parallel, and then concatenate the depth-aware multi-head self-attention (DA-MSA) outputs.
The process of DATL with LayerNorm(LN), multi-head \text{DA-SA1}, and \text{DA-SA2} at the current level is formulated as:
          \begin{equation}\widehat X^{l} = \text{DA-MSA1}(\text{LN}(\text{DA-MSA2}(\text{LN}(X^{l-1})))) + X^{l-1}\end{equation}
Then, an MLP module that has two fully-connected layers with GELU non-linearity between them is used for further feature transformations:
            \begin{equation}X^{l} = \text{MLP}(\text{LN}(\widehat X^{l}))) + \widehat X^{l}\end{equation}
Compared with global attention, local attention makes it possible for computation in high resolution.
\begin{table*}[ht!]
	\small
	\centering
	\resizebox{\textwidth}{!}{
	\begin{tabular}{c|c|cccccccc|c|cccccc}
		\hline
		\multirow{2}{*}{Methods}&
		\multicolumn{9}{c|}{Intermediate Group (F-score $\uparrow$)}&
		\multicolumn{7}{c}{Advanced Group (F-score $\uparrow$)} \\
		\cline{2-17}
		& Mean&Fam.&Fra.&Hor.&Lig.&M60&Pan.&Pla.&Tra. & Mean&Aud.&Bal.&Cou.&Mus.&Pal.&Tem. \\
		\hline
		MVSNet \cite{yao2018mvsnet} & 43.48&55.99&28.55&25.07&50.79&53.96&50.86&47.90&34.69 & -&-&-&-&-&-&- \\
		CasMVSNet \cite{gu2020cascade} & 56.84&76.37&58.45&46.26&55.81&56.11&54.06&58.18&49.51 & 31.12&19.81&38.46&29.10&43.87&27.36&28.11 \\
		UCS-Net \cite{conf/cvpr/ChengXZLLRS20} & 54.83&76.09&53.16&43.03&54.00&55.60&51.49&57.38&47.89 & -&-&-&-&-&-&- \\
		CVP-MVSNet \cite{conf/cvpr/YangMAL20} & 54.03&76.50&47.74&36.34&55.12&57.28&54.28&57.43&47.54 & -&-&-&-&-&-&- \\
		PVA-MVSNet \cite{conf/eccv/YiWDZCWT20} & 54.46&69.36&46.80&46.01&55.74&57.23&54.75&56.70&49.06 & -&-&-&-&-&-&- \\
		AA-RMVSNet \cite{wei2021aa} & 61.51&77.77&59.53&51.53&64.02&64.05&59.47&60.85&54.90 & 33.53&20.96&40.15&32.05&46.01&29.28&32.71 \\
		PatchmatchNet \cite{wang2021patchmatchnet} & 53.15&66.99&52.64&43.24&54.87&52.87&49.54&54.21&50.81 & 32.31&23.69&37.73&30.04&41.80&28.31&32.29 \\
		UniMVSNet \cite{unimvsnet} &64.36 &81.20 &\textbf{66.34} &53.11&\textbf{63.46}&66.09&64.84&\textbf{62.23}&57.53 & 38.96&28.33&44.36&39.74&52.89&33.80&34.63 \\
            \hline
            MVSTR \cite{Zhu2021MultiViewSW} &56.93 &76.92 &59.82 &50.16 &56.73 &56.53 &51.22 &56.58 &47.48 &32.85 &22.83 &39.04 &33.87 &45.46 &27.95 &27.97 \\
            
            TransMVS \cite{ding2022transmvsnet} & 63.52 &80.92 &65.83 &\textbf{56.94} &62.54 &63.06 &60.00 &60.20 &\textbf{58.67} &37.00 &24.84 &44.59 &34.77 &46.49 &\textbf{34.69} &36.62 \\
            MVSTER \cite{wang2022mvster} &- &- &- &- &- &- &- &- &- &37.53 &26.68 &42.14 &35.65 &49.37 &32.16 &\textbf{39.19}	\\
		\hline
		\textbf{CostFormer(PatchMatchNet)} & 56.27(+3.12)&72.46&52.59&54.27&	55.83&56.80&50.88&55.05&52.32 & 34.07(+1.76)&24.05&39.20&32.17&43.95&28.62&36.46 \\
		\textbf{CostFormer(PatchMatchNet*)} & 57.10(+3.95)&74.22&56.27&54.41&56.65&54.46&51.45&57.65&51.70 & 34.31(+2.00)&26.77&39.13&31.58&44.55&28.79&35.03 \\
		\textbf{CostFormer(UniMVSNet$^-$)} & 64.40(+0.04)&\textbf{81.45}&66.22&53.88&62.94&66.12&65.35&61.31&57.90 & \textbf{39.55(+0.59)}&28.61&\textbf{45.63}&\textbf{40.21}&52.81&34.40&35.62 \\
		\textbf{CostFormer(UniMVSNet*)} & \textbf{64.51(+0.15)}&81.31&65.51&55.57&\textbf{63.46}&\textbf{66.24}&\textbf{65.39}&61.27&57.30 & 39.43(+0.47)&\textbf{29.18}&45.21&39.88&\textbf{53.38}&34.07&34.87 \\
		\hline
	\end{tabular}}
         \caption{Quantitative results of different methods on the Tanks $\&$ Temples benchmark (higher is better). * is pretrained on DTU and fine-tuned on BlendedMVS. - is not pretrained on DTU and trained from scratch on BlendedMVS}
         \label{Quantitative-results-TT}
\end{table*}

However,  there is no connection across local windows with fixed partitions.
Therefore, regular and shifted window partitions are used alternately to enable cross-window connections. 
So at the next level, the window partition configuration is shifted along the height, width, and depth axes by $(\frac{h_{s}}{2}, \frac{w_{s}}{2}, \frac{d_{s}}{2})$. Depth-aware self-attention will be computed in these shifted windows(DAS-MSA); the whole process of DASTL can be formulated as:
           \begin{equation}\widehat X^{l+1} = \text{DAS-MSA1}(\text{LN}(\text{DAS-MSA2}(\text{LN}(X^{l})))) + X^{l}\end{equation}
           \begin{equation}X^{l+1} = \text{MLP}(\text{LN}(\widehat X^{l+1})) + \widehat X^{l+1}\end{equation}
\text{DAS-MSA1} and \text{DAS-MSA2} correspond to multi-head \text{Attention1} and \text{Attention2} within a shifted window, respectively.         
Assuming the number of stages is $n$, there are $n$ RDACT blocks in CostFormer. 

\subsection{Residual Regression Transformer}\label{RRT}
After aggregation, the cost $\widetilde{\mathcal{C}} \in \mathbb{R}^{HXWXD}$ will be used for depth regression.
To further explore the spatial relation under some depth, a Transformer block is applied to $\widetilde{\mathcal{C}}$ before softmax.
Inspired by the RDACT, the whole process of Residual Regression Transformer(RRT) can be formulated as:
                  \begin{equation}\widetilde{\mathcal{C}}_{k} = \text{RST}_{k}(\text{RT}_{k}(\widetilde{\mathcal{C}}_{k-1})), k=1,2,...,L\end{equation}
                  \begin{equation}\widetilde{\mathcal{ C}}_{out} = \text{RER}(\widetilde{\mathcal{C}}_{L}) + \widetilde{\mathcal{ C}}_{0}\end{equation}
where $\text{RT}_{k}$ is the k-th Regression Transformer layer with regular windows, $\text{RST}_{k}$ is the k-th Regression Transformer layer with shifted windows, RER is the re-embedding layer to recover the depth dimension from $\widetilde {\mathcal{C}}_{L}$, and it can be a 2D convolution with $D$ output channels.

$\text{RRT}$ also computes self-attention in a local window. 
Compared with $\text{RDACT}$, $\text{RRT}$ focuses more on spatial relations.
Compared with regular Swin \cite{liu2021swin} Transformer block, $\text{RRT}$ treats the depth as a channel, the number of channels is actually $1$ and this channel is squeezed before the Transformer. The embedding parameters are set to fit the cost aggregation of different iterations. If the embedding dimension number equals $D$, $\widetilde{\mathcal{C}}_{out}$ can be simply formulated as: 
                 \begin{equation}\widetilde{\mathcal{C}}_{out} = \widetilde{\mathcal{C}}_{L} + \widetilde{\mathcal{C}}_{0}\end{equation}
As a stage may iterate many times with different depth hypotheses, the number of RRT blocks should be set the same as the number of iterations. The whole $\text{RRT}$ is shown in the yellow window in Figure \ref{fig:CostFormer}. 

\section{Training}

\subsection{Loss function}
Final loss combines with the losses of all iterations at all stages and the loss from the final refinement module:
       \begin{equation}Loss = \overset{s}{\underset{k=1}{\sum}} \overset{n}{\underset{i=1}{\sum}}L_{i}^{k} + L_{ref}\end{equation}
where $L_{i}^{k}$ is the regression or unification loss of the $i$-th iteration at $k$-th stage.
$L_{ref}$ is the regression or unification loss from refinement module. If refinement module does not exist, the $L_{ref}$ loss is set to zero.

\subsection{Common training settings}
CostFormer is implemented by Pytorch \cite{NEURIPS2019_9015}. For RDACT, we set the depth number at stages 3, 2, 1 as 4, 2, 2; patch size at height, width and depth axes as 4, 4, 1;  window size at height, width and depth axes as 7, 7, 2. If the backbone is set as PatchMatchNet, embedding dimension number at stages 3, 2, 1 are set as 8, 8, 4.
For RRT, we set the depth number as 2 at all stages, patch size as 1 at all axes; window size as 8 at all axes. If the backbone is set as PatchMatchNet, embedding dimension number at iteration 2, 2, 1 at stages 3, 2, 1 as 32, 64, 16, 16, 8. All models are trained on Nvidia GTX V100 GPUs.
After depth estimation, we reconstruct point clouds similar to MVSNet \cite{yao2018mvsnet}.

\section{Experiments}
\label{sec_experiments}
In this section, we introduce multiple MVS datasets and evaluate our method on these datasets.
The results will be further reported in detail. 
\subsection{DATASETS}
The datasets used in the evaluation are DTU \cite{journals/ijcv/AanaesJVTD16}, BlendedMVS \cite{yao2020blendedmvs}, ETH3D \cite{conf/cvpr/SchopsSGSSPG17}, Tanks $\&$ Temples \cite{journals/tog/KnapitschPZK17}, and YFCC-100M \cite{journals/cacm/ThomeeSFENPBL16}.
The DTU dataset is an indoor multi-view stereo dataset with 124 different scenes, there are 49 views under seven different lighting conditions in one scene. 
Tanks $\&$ Temples is collected in a more complex and realistic environment, and it’s divided into the intermediate and advanced set.
ETH3D benchmark consists of calibrated high-resolution images of scenes with strong viewpoint variations.
It is divided into training and test datasets.
While the training dataset contains 13 scenes,  the test dataset contains 12 scenes.
BlendedMVS dataset is a large-scale synthetic dataset, consisting of 113 indoor and outdoor scenes and split into 106 training scenes and 7 validation scenes.
\begin{table}[t!]
\centering
\tiny
\resizebox{\linewidth}{!}{
\begin{tabular}{l|l|l|l}
    \hline
    Methods   & Acc. (mm) & Comp. (mm) & Overall (mm) \\
    \hline
    Furu \cite{journals/pami/FurukawaP10}    & 0.613       & 0.941    & 0.777 \\
    Tola \cite{journals/mva/TolaSF12}    & 0.342    & 1.190    & 0.766 \\
    Gipuma \cite{conf/iccv/GallianiLS15}    & \textbf{0.283}    & 0.873    & 0.578 \\
    Colmap \cite{conf/cvpr/SchonbergerF16}    & 0.400    & 0.644    & 0.532 \\
    \hline
    SurfaceNet \cite{conf/iccv/JiGZLF17}    & 0.450    & 1.040    & 0.745 \\
    MVSNet \cite{yao2018mvsnet}    & 0.396   & 0.527    & 0.462 \\
    R-MVSNet \cite{conf/cvpr/0008LLSFQ19}    & 0.383   & 0.452   & 0.417 \\
    P-MVSNet \cite{conf/iccv/LuoGJHL19}    & 0.406   & 0.434   & 0.420 \\
    Point-MVSNet \cite{conf/iccv/ChenHXS19}    & 0.342   & 0.411   & 0.376 \\
    Fast-MVSNet \cite{conf/cvpr/YuG20}    & 0.336   & 0.403   & 0.370 \\
    CasMVSNet \cite{gu2020cascade}    & 0.325   & 0.385   & 0.355 \\
    UCS-Net \cite{conf/cvpr/ChengXZLLRS20}    & 0.338   & 0.349  & 0.344 \\
    CVP-MVSNet \cite{conf/cvpr/YangMAL20} & 0.296   & 0.406  & 0.351 \\
    PVA-MVSNet \cite{conf/eccv/YiWDZCWT20} & 0.379  &0.336 &0.357 \\
    PatchMatchNet \cite{wang2021patchmatchnet} & 0.427   & 0.277  & 0.352 \\
    AA-RMVSNet \cite{wei2021aa} & 0.376   & 0.339  & 0.357 \\
    UniMVSNet \cite{unimvsnet} &0.352 &0.278 &0.315 \\
    \hline
    \textbf{CostFormer(Based on PatchMatchNet)} &0.424 &\textbf{0.262} &0.343 (+0.0093) \\
    \textbf{CostFormer(Based on CasMVSNet)} &0.378 &0.313 &0.345 \textbf{(+0.0097)} \\
    \textbf{COstFormer(Based on UniMVSNet)} &0.301 &0.322 &\textbf{0.312} (+0.0035) \\
    \hline    
\end{tabular}}
    \caption{Quantitative results of different methods on DTU.} 
    \label{Quantitative-results-DTU}
\end{table}
\subsection{Main Settings and Results on DTU}
For the evaluation on the DTU \cite{journals/ijcv/AanaesJVTD16} evaluation set, we only use the DTU training set. During the training phase, we set the image resolution to 640 × 512. We compare our method to recent learning-based MVS methods, including CasMVSNet \cite{gu2020cascade} and PatchMatchNet \cite{wang2021patchmatchnet} which are also set as backbones of CostFormer. We follow the evaluation metrics provided by the DTU dataset. The quantitative results on the DTU evaluation set are summarized in Table \ref{Quantitative-results-DTU}, which indicates that the plug-and-play CostFormer improves the cost aggregation. Partial visualization results of Table \ref{Quantitative-results-DTU} are shown in Figure \ref{fig:dtu_comparison}.


\noindent \textbf{Complexity Analysis}: For the complexity analysis of CostFormer, we plug it into PatchMatchNet \cite{wang2021patchmatchnet} and first compare the memory consumption and run-time with this backbone. For a fair comparison, a fixed input size of 1152 × 864 is used to evaluate the computational cost on a single GPU of NVIDIA Telsa V100. Memory consumption and run-time of PatchMatchNet \cite{wang2021patchmatchnet} are 2323MB and 0.169s. They are only increased to 2693MB and 0.231s by the plug-in. 


Based on the reports of PatchMatchNet \cite{wang2021patchmatchnet}, we then get the comparison results of other state-of-the-art learning-based methods. Memory consumption and run-time are reduced by 61.9$\%$ and 54.8$\%$ compared to CasMVSNet \cite{gu2020cascade}, by 48.8$\%$ and 50.7$\%$ compared to UCSNet \cite{conf/cvpr/ChengXZLLRS20} and by 63.5$\%$ and 77.3$\%$ compared to CVP-MVSNet \cite{conf/cvpr/YangMAL20}.
Combining the results(lower is better) are shown in Table \ref{Comparison-GPU-memory} and Figure \ref{fig:compar}, GPU memory and run-time of CostFormer are set as 100$\%$.

\begin{table}[h]
\centering
\tiny
\resizebox{\linewidth}{!}{
\begin{tabular}{l|l|l|l}
    \hline
    Method           & GPU Memory ($\%$) & Run-time ($\%$) & Overall (mm) \\
    \hline
    CasMVSNet \cite{gu2020cascade}    & 262.47$\%$           & 221.24$\%$      & 0.355       \\
    UCSNet \cite{conf/cvpr/ChengXZLLRS20}    & 195.31$\%$           & 202.84$\%$       & 0.344       \\
    CVP-MVSNet \cite{conf/cvpr/YangMAL20}    & 273.97$\%$           & 440.53$\%$       & 0.351       \\
    \textbf{Ours} & 100.00$\%$          & 100.00$\%$       & 0.343      \\
    \hline     
\end{tabular}}
    \caption{ Comparison with other SOTA learning-based MVS methods on DTU. Relationship between overall performance, GPU memory and run-time.}
    \label{Comparison-GPU-memory}
\end{table}

\noindent \textbf{Comparison with Transformers} We also compare CostFormer with other Transformers \cite{Zhu2021MultiViewSW,wang2022mvster,ding2021transmvsnet,liao2022wtmvsnet} which are used in MVS methods and not plug-and-play. For a fair comparison, only direct improvements(higer is better) and incremental cost of run time(low is better) from pure Transformers under similar depth hypotheses are summarized in Table \ref{comprable-trans}. 
\begin{table}[h!]
\tiny
\resizebox{\linewidth}{!}{
\begin{tabular}{l|l|l|l}
\hline
Method      & Trans Improvement (mm) &Delta Time (s)   &Delta Time (\%) \\
\hline
MVSTR \cite{Zhu2021MultiViewSW}       & +0.0140    &+0.359s & +78.21\%  \\
\hline
TransMVS \cite{ding2021transmvsnet}    & +0.0160    &+0.367s  & +135.42\% \\
\hline
WT-MVSNet(CT) \cite{liao2022wtmvsnet} &+0.0130 &+0.265s &- \\
\hline
MVSTER(CNN Fusion) \cite{wang2022mvster} & +0.0040    &+0.016s     & +13.34\% \\
\hline
CostFormer(CNN Fusion)  & +0.0097  & +0.062s  & +36.69\%  \\
\hline
\end{tabular}}
\caption{Quantitative improvement of performance and incremental
 cost of run time of different Transformers on DTU evaluation set.}
 \label{comprable-trans}
\end{table}


\subsection{Main Settings and Results on Tanks $\&$ Temples}
For the evaluation on Tanks $\&$ Temples \cite{journals/tog/KnapitschPZK17}, we use the DTU \cite{journals/ijcv/AanaesJVTD16} dataset and the Blended MVS \cite{yao2020blendedmvs} dataset.
We compare our method to those recent learning-based MVS methods, including PatchMatchNet \cite{wang2021patchmatchnet} and UniMVSNet \cite{unimvsnet} which are also set as backbones of CostFormer.
The quantitative results on the Tanks $\&$ Temples \cite{journals/tog/KnapitschPZK17} set are summarized in Table \ref{Quantitative-results-TT}, which indicates the robustness of CostFormer. Partial visualization results of Table \ref{Quantitative-results-TT} are shown in Figure \ref{fig:tanks_comparison}. We would like to clarify that UniMVSNet$^-$ in Table \ref{Quantitative-results-TT} only uses BlendedMVS for training which uses less data (no DTU) than the UniMVSNet baseline.

\subsection{Main Settings and Results on ETH3D} 
We use the PatchMatchNet \cite{wang2021patchmatchnet} as backbone and adopt the trained model used in the Tanks $\&$ Temples dataset \cite{journals/tog/KnapitschPZK17} to evaluate the ETH3D \cite{conf/cvpr/SchopsSGSSPG17} dataset. As shown in Table \ref{ETH3D-result}, our method outperforms others on both the training and particularly challenging test datasets(higher is better). 

\begin{table}[htp]
	\tiny
	\resizebox{\linewidth}{!}{
	\begin{tabular}{l|ll|ll}
		\hline
		\multicolumn{1}{c|}{\multirow{2}{*}{Methods}} & \multicolumn{2}{c|}{Training}                          & \multicolumn{2}{c}{Testing}                              \\ \cline{2-5} 
		\multicolumn{1}{c|}{}                         & \multicolumn{1}{l|}{F1 score $\uparrow$} & Time(s) $\downarrow$               & \multicolumn{1}{l|}{F1 score  $\uparrow$} & Time(s) $\downarrow$               \\ \hline
		MVE \cite{conf/vast/FuhrmannLG14}                                            & \multicolumn{1}{l|}{20.47}    & 13278.69               & \multicolumn{1}{l|}{30.37}    & 10550.67               \\ \hline
		Gipuma \cite{conf/iccv/GallianiLS15}                                         & \multicolumn{1}{l|}{36.38}    & 587.77                 & \multicolumn{1}{l|}{45.18}    & 689.75                 \\ \hline
		PMVS \cite{journals/pami/FurukawaP10}                                           & \multicolumn{1}{l|}{46.06}    & 836.66                 & \multicolumn{1}{l|}{44.16}    & 957.08                 \\ \hline
		COLMAP \cite{conf/cvpr/SchonbergerF16}                                         & \multicolumn{1}{l|}{67.66}    & 2690.62                & \multicolumn{1}{l|}{73.01}    & 1658.33                \\ \hline
		PVSNet \cite{journals/corr/abs-2007-07714}                                         & \multicolumn{1}{l|}{67.48}    & \multicolumn{1}{c|}{-} & \multicolumn{1}{l|}{72.08}    & 829.5                  \\ \hline
            IterMVS \cite{wang2021itermvs}  &  \multicolumn{1}{l|}{66.36}  &  \multicolumn{1}{c|}{-}  &  \multicolumn{1}{l|}{74.29}  &  \multicolumn{1}{c}{-}  \\ \hline
		PatchMatchNet \cite{wang2021patchmatchnet}                                  & \multicolumn{1}{l|}{64.21}    & 452.63                 & \multicolumn{1}{l|}{73.12}    & 492.52                 \\ \hline
		PatchMatch-RL \cite{lee2021patchmatchrl}                                  & \multicolumn{1}{l|}{67.78}    & \multicolumn{1}{c|}{-} & \multicolumn{1}{l|}{72.38}    & \multicolumn{1}{c}{-}  \\ \hline
		\textbf{CostFormer(Ours)}                                           & \multicolumn{1}{l|}{\textbf{68.92(+4.71)}}    & 566.18 & \multicolumn{1}{l|}{\textbf{75.24(+2.12)}}    & 547.64 \\ \hline
	\end{tabular}}
       \caption{Quantitative results of different methods on ETH3D.}
       \label{ETH3D-result}
\end{table}

\subsection{Main Settings and Results on BlendedMVS dataset}
We use the model used in ETH3D. On BlendedMVS \cite{yao2020blendedmvs} evaluation set, we set $N=5$ and image resolution as 576 × 768.  End point error (EPE), 1 pixel error (e1), and 3 pexels error (e3) are used as the evaluation
metrics. Quantitative results(lower is better) of different methods are shown in Table \ref{Blended-presult}.   
\begin{table}[h!]
\centering
\tiny  
\resizebox{0.9\linewidth}{!}{
\begin{tabular}{l|l|l|l}
\hline
Method                    & EPE           & e1 (\%)         & e3 (\%)        \\ \hline
MVSNet \cite{yao2018mvsnet}                    & 1.49          & 21.98          & 8.32          \\ \hline
MVSNet-s \cite{DBLP:journals/corr/abs-2104-15119}                  & 1.35          & 25.91          & 8.55          \\ \hline
CVP-MVSNet \cite{yang2020cost}                & 1.90          & 19.73          & 10.24         \\ \hline
VisMVSNet \cite{zhang2020visibility}                 & 1.47          & 18.47          & 7.59          \\ \hline
CasMVSNet \cite{gu2020cascade}                 & 1.98          & 15.25          & 7.60          \\ \hline
EPPMVSNet \cite{ma2021epp}  &  1.17  &  12.66  &  6.20  \\ \hline
TransMVSNet  \cite{ding2021transmvsnet}  &  0.73   &  8.32  &  3.62  \\ \hline
\textbf{CostFormer(Based on PatchmatchNet)} & {0.84} & {12.37} & {4.59} \\ \hline
\textbf{CostFormer(Based on UniMVSNet)} & \textbf{0.43} & \textbf{7.05} & \textbf{2.70} \\ \hline
\end{tabular}}
\caption{Quantitative results of different methods on BlendedMVS}
\label{Blended-presult}
\end{table}

\section{Conclusion}
\label{sec_conclusion}
In this work, we explore whether cost Transformer can improve the cost aggregation and propose a novel CostFormer with the cascade RDACT and RRT modules. The experimental results on DTU \cite{journals/ijcv/AanaesJVTD16} , Tanks $\&$ Temples  \cite{journals/tog/KnapitschPZK17}, ETH3D \cite{conf/cvpr/SchopsSGSSPG17}, and BlendedMVS \cite{yao2020blendedmvs} show that our method is competitive, efficient, and plug-and-play.
Cost Transformer can be your need for better cost aggregation in multi-view stereo.

\bibliographystyle{named}
\bibliography{ijcai23}

\end{document}